\documentclass[sigconf]{acmart}

\AtBeginDocument{%
  \providecommand\BibTeX{{%
    \normalfont B\kern-0.5em{\scshape i\kern-0.25em b}\kern-0.8em\TeX}}}

\acmPrice{15.00}
\acmISBN{978-1-4503-XXXX-X/18/06}

\usepackage{amsmath}
\usepackage{wrapfig}
\usepackage{booktabs} 
\usepackage{array}  %
\usepackage{multirow}
\usepackage{bm}
\usepackage[ruled,vlined]{algorithm2e}
\usepackage[utf8]{inputenc} 
\usepackage{enumitem}
\usepackage{url}            
\usepackage{booktabs}       
\usepackage{amsfonts}       
\usepackage{nicefrac}       
\usepackage{microtype}      
\usepackage{xcolor}         
\usepackage{amsmath}
\usepackage{dsfont}
\usepackage{graphicx}
\usepackage{balance}
\usepackage{float}
\usepackage[capitalize]{cleveref}
\crefname{section}{Sec.}{Secs.}
\Crefname{section}{Section}{Sections}
\Crefname{table}{Table}{Tables}
\crefname{table}{Tab.}{Tabs.}

\copyrightyear{2022}
\acmYear{2022}
\setcopyright{acmlicensed}
\acmConference[MM '22]{Proceedings of the 30th ACM International Conference on Multimedia}{October 10--14, 2022}{Lisboa, Portugal}
\acmBooktitle{Proceedings of the 30th ACM International Conference on Multimedia (MM '22), October 10--14, 2022, Lisboa, Portugal}
\acmPrice{15.00}
\acmDOI{10.1145/3503161.3548395}
\acmISBN{978-1-4503-9203-7/22/10}

\begin{document}

\title{DeViT: Deformed Vision Transformers in Video Inpainting}


\author{Jiayin Cai}

\affiliation{%
  \institution{Kuaishou Technology}
}
\email{caijiayin@kuaishou.com} 

\author{Changlin Li}
\affiliation{%
  \institution{Kuaishou Technology}
    }
\email{lichanglin@kuaishou.com}

\author{Xin Tao}
\affiliation{%
  \institution{Kuaishou Technology}
}
\email{taoxin@kuaishou.com}

\author{Chun Yuan}
\authornote{Corresponding authors. yuanc@sz.tsinghua.edu.cn, yuwing@gmail.com\vspace{-0.1in}}
\affiliation{%
  \institution{SIGS, Tsinghua University}
  \institution{Peng Cheng National Laboratory}
}
\email{yuanc@sz.tsinghua.edu.cn}

\author{Yu-Wing Tai}
\authornotemark[1]
\affiliation{%
  \institution{Kuaishou Technology}
}
\email{yuwing@gmail.com}


\renewcommand{\shortauthors}{Jiayin Cai et al.}
\begin{abstract}
This paper proposes a novel video inpainting method. We make three main contributions: 
First, we extended previous Transformers with patch alignment by introducing Deformed Patch-based Homography (DePtH), which improves patch-level feature alignments without additional supervision and benefits challenging scenes with various deformation.
Second, we introduce Mask Pruning-based Patch Attention (MPPA) to improve patch-wised feature matching by pruning out less essential features and using saliency map. MPPA enhances matching accuracy between warped tokens with invalid pixels.
Third, we introduce a Spatial-Temporal weighting Adaptor (STA) module to obtain accurate attention to spatial-temporal tokens under the guidance of the Deformation Factor learned from DePtH, especially for videos with agile motions. 
Experimental results demonstrate that our method outperforms recent methods qualitatively and quantitatively and achieves a new state-of-the-art.
\vspace{-0.2in}
\end{abstract}

\ccsdesc[500]{Theory of computation~Design and analysis of algorithms}
\ccsdesc[500]{Computing methodologies~Computer vision problems}
\keywords{Video Inpainting, Transformer Networks, Patch-based Homography Estimator, Mask Pruning-based Patch Attention
\vspace{-0.05in}}



\begin{teaserfigure}
  \vspace{-0.1in}
  \includegraphics[width=\textwidth]{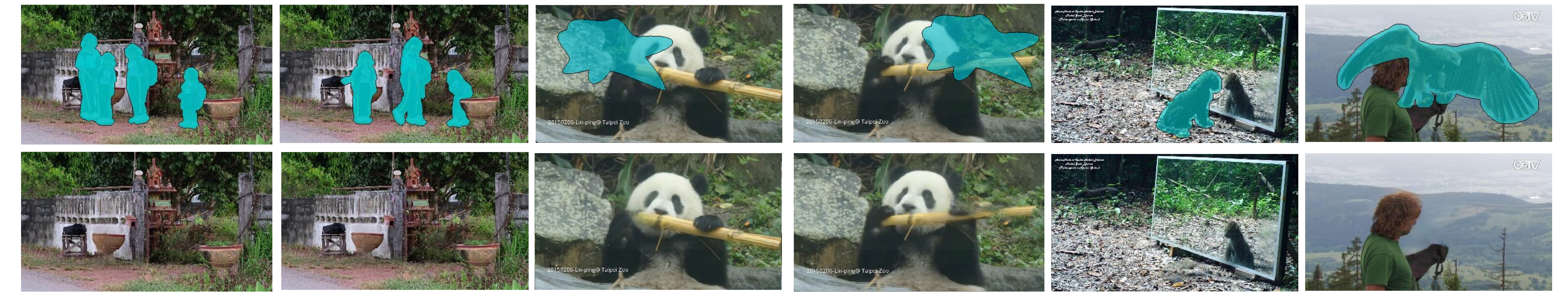}
  \vspace{-0.25in}
  \caption{\label{fig1} Video inpainting results generated by our proposed model. The results show high capability for recovering texture detail from spatial structure for slow panning motions, and temporal coherent for agile motion. Best viewed with zoom-in.}
  \Description{Enjoying the baseball game from the third-base
  seats. Ichiro Suzuki preparing to bat.}
  \label{fig:teaser}
\end{teaserfigure}

\maketitle

\vspace{-0.15in}
\section{Introduction}
\label{sec:intro}
Video inpainting aims to replace unwanted objects or scratches in a video with plausible contents. Since different frames in a video often have spatial misalignment and structure distortion, it is very challenging to recover both spatially plausible and temporally consistent results. Thus, the key to video inpainting is effectively establishing correspondences between target patches with holes and distorted reference patches. Previous learning-based approaches adopt either patch matching (patch-match~\cite{wexler2004space,newson2014video}, patch-based attention~\cite{woo2019align,pathak2016context}) or pixel-wise alignment~\cite{Copy-and-paste} (optical ﬂow~\cite{Deep-flow-guided,li2020short,gao2020flow}, deformable convolution~\cite{hara2021video}). While many works have been proposed, their results are still far from perfect. This paper investigates three key challenges in video inpainting and proposes corresponding remedies to tackle each one of them under a unified framework:

\textbf{Alignment in Video Inpainting}
Many previous methods estimate global alignment~\cite{Copy-and-paste} or optical flow~\cite{Deep-flow-guided,gao2020flow} merely based on known regions and propagate valuable contents from external frames into missing areas. However, they usually fail when faced with occluded stationary background, where there are no valuable hints for hole areas. On the other hand, recent patch-based methods~\cite{darabi2012image,simakov2008summarizing} may also fail to obtain high-quality source patches due to geometric variation in object scale, pose, viewpoint or part deformation. Thus it requires network modules to accommodate all these variations.

To relieve the limitations above, we propose a \emph{Deformed Patch-based Transformer Homography} (DePtH) module, which aligns key/value components of the transformer to the query, in a patch-wised way, before calculating feature similarity as in vanilla Vision Transformer.
Our proposed DePtH robustly warps spatially misaligned patches, which can produce a more precise soft attention map and better-aligned candidate patches, thus yielding better inpainting results.


\textbf{Feature Matching with Invalid Feature}
 Another critical challenge in video inpainting lies in effectively establishing correspondences between hole patches and complete patches. Previous patch-wised matching methods search for related patches by calculating feature similarity~\cite{STTN, liu2021decoupled, pathak2016context, woo2019align}, (usually Scaled Dot-Product~\cite{vaswani2017attention} in attention mechanism). However, the calculation of such similarity is easily distracted by hole pixels. The features of the hole pixel will participate in the matching score calculation,~\emph{e.g.}, two patches with holes in the same position usually produce a high similarity score, which prevents the process from locating high-quality semantically similar candidates. Previous work~\cite{STTN, liu2021decoupled} adopt a straightforward way to set patches-pair scores to zero if the ratio of the invalid pixel is larger than 0.5. Such computation is too coarse to precisely model similarity between them.

To overcome the shortcomings of the vanilla attention mechanism, we design a new operator \emph{Mask Pruning based Patch Attention} (MPPA) with a new hole-aware similarity measure on patch-level. MPPA aims to achieve a better elaborate matching among the tokens containing invalid pixels. Therefore, the most relevant patches can be detected and prevent the model from generating blurry results influenced by invalid pixels.

\textbf{Spatial-Temporal Weighting Adaptor}
Video inpainting methods usually borrow valid contents from neighboring frames~\cite{nazeri2019edgeconnect, Onion-peel}, where missing contents may appear. However, when scenes in the video are almost static, those methods tend to produce blurry results with few details. In this case, the problem degrades to a single image inpainting task. We, in this paper, introduce a novel weighting adaptive strategy so that the network will adaptively deal with significant transformations for temporal object movement during different frames as well as stationary background texture in one frame.
To summarize, our main contributions are as follows: 
\begin{itemize}[leftmargin=*,nosep,nolistsep]
    \item We propose a new network \emph{Deformed Vision Transformer} (DeViT) with emphasis on better patch-wise alignment and matching in video inpainting. The results are promising and superior to the current state-of-the-art on common inpainting test datasets.
    
    \item We introduce patch-based homography into the attention module of the transformer, and propose a novel \emph{Deformed Patch based Homography} (DePtH), which is designed to estimate alignment parameters for patches, and learn a Deformation Factor for guiding STA. Aligned tokens are then adaptively selected by MPPA, which could improve feature matching quality.
    
    \item A new patch similarity measurement considering correlation map between patches has been proposed in \emph{Mask Pruning based Patch Attention} (MPPA), so the most relevant patches can be located and prevent the model from generating blurry results influenced by invalid pixels.
    
    \item \emph{Spatial Temporal weighting Adaptor} (STA) is proposed to adaptively attend to different spatial-temporal tokens more accurately under the guidance of the Deformation Factor, indicating the motion type of the video. It can better handle scenes like complex motion as well as scenes with a stationary background.
\end{itemize}

\vspace{-6mm}
\section{Related Work}
\label{Related Work}

\textbf{Traditional Inpainting Methods} Early works for image inpainting can broadly fall into either diffusion-based~\cite{ballester2001filling,levin2003learning} or patch-based methods~\cite{darabi2012image}. The former propagates texture from the hole boundaries towards the hole center, and works well with small holes, but suffers artifacts and noisy results with large holes. The latter tries to match and copy the nearest neighbor background patches, and is widely deployed in practical applications.

\textbf{Learning based Video Inpainting Methods} Wang ~\cite{wang2018perceptual} proposed the first deep learning-based video inpainting using a 3D encoder-decoder network. However, this work did not cover the object removal task in general videos and was only applied to a few specific domains. In order to complete the missing contents in the video, Kim ~\cite{kim2019deep} proposed a 3D-2D encoder-decoder network to maintain temporal consistency through recurrent feedback and a memory layer with the flow and the warping loss. Wang ~\cite{wang2019video} used a 3D CNN to capture the temporal structure from low-resolution video and a 2D CNN to capture spatial details. Nevertheless, their fundamental limitation is that the temporal window for the referencing is small. Therefore, using valid pixels in distant frames is difficult, resulting in a limited performance for scenes with large or slow-moving objects.

\begin{figure*}[t]
  \centering
  \includegraphics[width=0.98\textwidth]{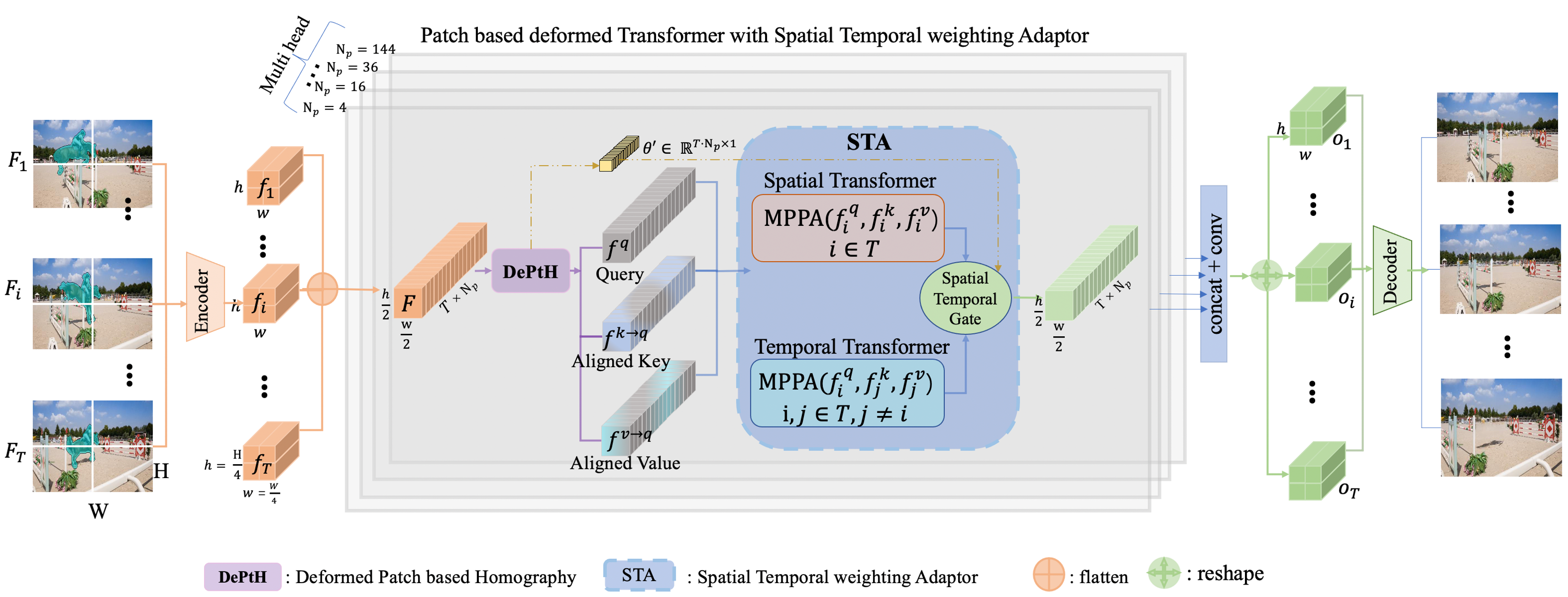}
  \caption{The overview of our framework. DeViT consists of 1) a frame-level encoder and decoder, 2) Deformed Patch based Homography (DePtH) to warp and align each patched-wise Key (K) and Value (V) to Query (Q) and 3) A Spatial-temporal Transformer with weighting Adaptor (STA), which basic attention operator is MPPA. More details can be found in section \ref{Method}.}
  \label{fig2}
\end{figure*}

\textbf{Alignment Methods}
Granados  ~\cite{granados2012background} proposed to align the frames based on the homographies. They also applied the optical flow between completed frames to maintain temporal consistency. Woo ~\cite{Copy-and-paste} proposed a deep alignment network to compute affine matrices between images to align frames with affine matrices. Jaderberg  ~\cite{STMjaderberg2015spatial} proposed to use a spatial transformer that performs explicit spatial transformations of features. Dai ~\cite{dai2017deformable} proposed deformable convolution to enhance CNNs' capability of modeling geometric transformations. Newson ~\cite{newson2014video} proposed 3D Patch Match to maintain the temporal consistency rather than using the affine transformation to compensate for the motion. Huang ~\cite{huang2016temporally} proposed using optical flow optimization in spatial patches to complete images while preserving temporal consistency. Woo ~\cite{woo2019align} used homography-based alignment instead of flow-guided warping. However, all the methods above are based on frame level, which can not realize elaborative alignment for complex movement in the video.

\textbf{Patch-wise Attention} Attention mechanisms so far have been designed to look at the same patterns in parts of the images (patches) across different frames in a video, which can predict the best location to attend to in the frames. Oh ~\cite{Onion-peel} proposes asymmetric attention for calculating the similarities between hole regions on the target frame and valid regions on reference frames. Zeng ~\cite{STTN} proposed applying a transformer to video inpainting tasks to match and retrieve information from multiple frames with infinite space-time receptive fields. However, this method can not deal with the misalignment in orientation and scale conversion between patches, suffering from blurry outputs by averaging unaligned reference patches. Liu ~\cite{liu2021decoupled} propose a decoupled spatial-temporal transformer, which calculates the attention between tokens from different frames' same zone in the temporal transformer. However, it can not handle non-rigid transformation and agile/complex motion, especially for largely misaligned references. So we propose MPPA in this model, which pruned out invalid pixels in warped tokens first and then wight matching scores with saliency map.

\vspace{-2mm}
\section{Method}
\label{Method}
Our framework takes a set of video frames $\mathcal{X}:=\left\{X_{1}, X_{2}, \ldots, X_{T}\right\}$ and hole masks $\mathcal{M}:=\left\{M_{1}, M_{2} as input, \ldots, M_{T}\right\}$, where $M_{t}$ is a one-channel binary matrix where 1 represent pixels in holes and 0 represent valid pixels outside the hole. 
And we denote $\mathcal{Y}:=\left\{Y_{1}, Y_{2}, \ldots, Y_{T}\right\}$ as the ground truth frames. We aim to learn a mapping $ G: \mathcal{X} \rightarrow \mathcal{Y}$ such that the prediction $\hat{\mathcal{Y}}$ be as close as possible to the ground truth video $\mathcal{Y}$, while being plausible and consistent in space and time. 

\vspace{-4mm}
\subsection{Overall Design}

\textbf{Whole Architecture} The framework overview is shown in Figure~\ref{fig2}. Specifically, our network consists of three components: a frame-level encoder and decoder, a Deformed Patch-based Homography (DePtH), and a Spatial-temporal Transformer with weighting adaptor (STA), which basic attention operator is MPPA. The first core component, DePtH, warps and aligns each patched-wise Key (K) and Value (V) to Query (Q). Q, aligned K, and aligned V, serve as the basic vector in our transformer, are sent to MPPA, a generalization of the conventional attention operator, to conduct aligned-patch-based pruning attention. STA is adopted in our spatial-temporal transformer to adaptively allocate different weights to spacial and temporal branches for different videos with variance moving speed. 

\textbf{Network Design} The video is processed frame-by-frame in the temporal order. In order to fill a target frame $X_{t}$, our network takes  $X_{t-n}^{t+n}$, neighboring frames around $t$-{th} frame and $X_{1, s}^{T}$, distant frames with interval $s$, then complete all the input frames simultaneously. Where $t$ is the target frame index, $n$ is neighbour window size. 

\textbf{Image Encoder} The image encoder takes an image of $240\times 432$ pixels as input, and produces an embedded feature map in $\mathds{R}^{c \times 60 \times 108}$, where $c$ denotes the channel size. We use a shared weight encoder for all frames and obtain $f_{t}$ for $t$-th frame. We extract $N_{p}=n^{2}$ patches from each frame where $N_{p}$ is number of patches in each frame, which is different in multi head (we use $n=2,3,6,12$ or $N_{p}=4,9,36,144$ in our experiment). Different heads of the transformer attends on patches across different scales. In Figure~\ref{fig2}, we show the case of 4 patches for each frame.
 Then we concat these $N_{p}$ patches of $T$ frames as patch features $F:\mathds{R}^{N \times c \times h \times w}$. The set of patched feature $F$ will be sent to Deformed Patch Based Homography as the input, where $N=T*N_{p}$ is the whole number of patches in the video.


\vspace{-4mm}
\subsection{Deformed Patch based Homography}

\begin{figure}[t]
  \begin{center}
  \vspace{-4mm}
    \includegraphics[width=0.45\textwidth]{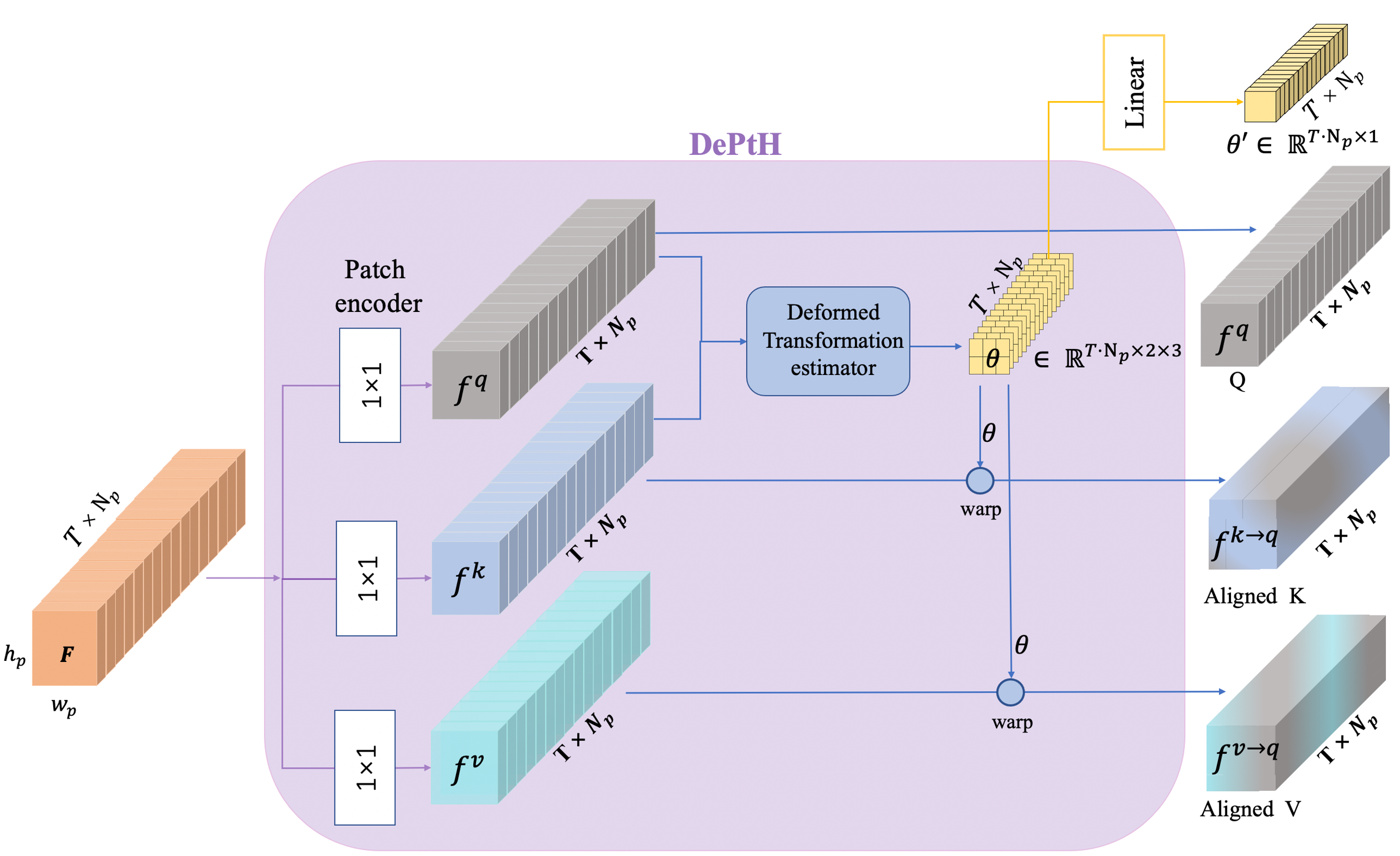}
    \caption{Deformed Patch based Homography(DePtH). DePtH is designed to estimate alignment parameters $\theta$ for patches pair, and learn a Deformation Factor $\theta'$ for guiding STA. Each patch is aligned independently with the estimated affine matrix $\theta$ to get the aligned Key token $\boldsymbol{f}^{k \rightarrow q}$ and Value token $\boldsymbol{f}^{v \rightarrow q}$.}
    \vspace{-4mm}
  \end{center}
\label{fig:DePtH}
\end{figure}

Traditional Transformers are inherently limited to the complex scenario or agile motions. The limitation originates from the fixed geometric structures of Transformer modules: patches are sampled from feature maps at fixed locations and fixed shapes. Attention operations are conducted from shape-fixed patches from query key and value, which limits the precision and accuracy of patch matching. Another limitation is that these attention-based methods usually assume global affine transformations or homogeneous motions, making them hard to model complex motions and often yield blurry edges in detail. To tackle these problems, We apply a Deformed Patch-based Homography(DePtH) into a multi-head spatial-temporal transformer:

\textbf{Patch Encoder} In the embedding step, we encode the output of image encoder into query, key and value for further retrieval: $ \boldsymbol{f}^{q}, \boldsymbol{f}^{k}, \boldsymbol{f}^{v}\footnote{All the superscript $q$, $k$, $v$ in our paper indicate query, key, and value component in transformer.}=D_{q}\left(F\right), D_{k}\left(F\right), D_{v}\left(F\right)$, where $D_{q}(\cdot), D_{k}(\cdot) \text{ and } D_{v}(\cdot)$ denote the $1\times1$ 2D convolutions that embed input features into query, key and value feature spaces while maintaining the spatial size of features.

\textbf{Deformed Patch-based Transformation Estimator}
Given a set of patched features, transformation estimator could predict Deformation Factor $\theta$, which is to warp and align each patched key feature and patched value feature onto the patched query feature. The transformation estimator takes the patch-based Query feature map and patch-based K feature map as input and produces homography parameters $\theta$ between the reference and target. It is trained to output 6 parameters (affine transformation): $\boldsymbol{T}(\boldsymbol{f}^{q},\boldsymbol{f}^{k}) =\theta$, and $\boldsymbol{T}: \mathds{R}^{N\times 2c \times wh} \rightarrow \mathds{R}^{N \times 2\times3}$. Note that the DePtH is jointly trained with other networks in an end-to-end manner, not independently. Details of the Deformed Transformation estimator network architectures are provided in the supplementary materials.

\begin{figure}[t]
  \begin{center}
  \vspace{-4mm}
    \includegraphics[width=0.45\textwidth]{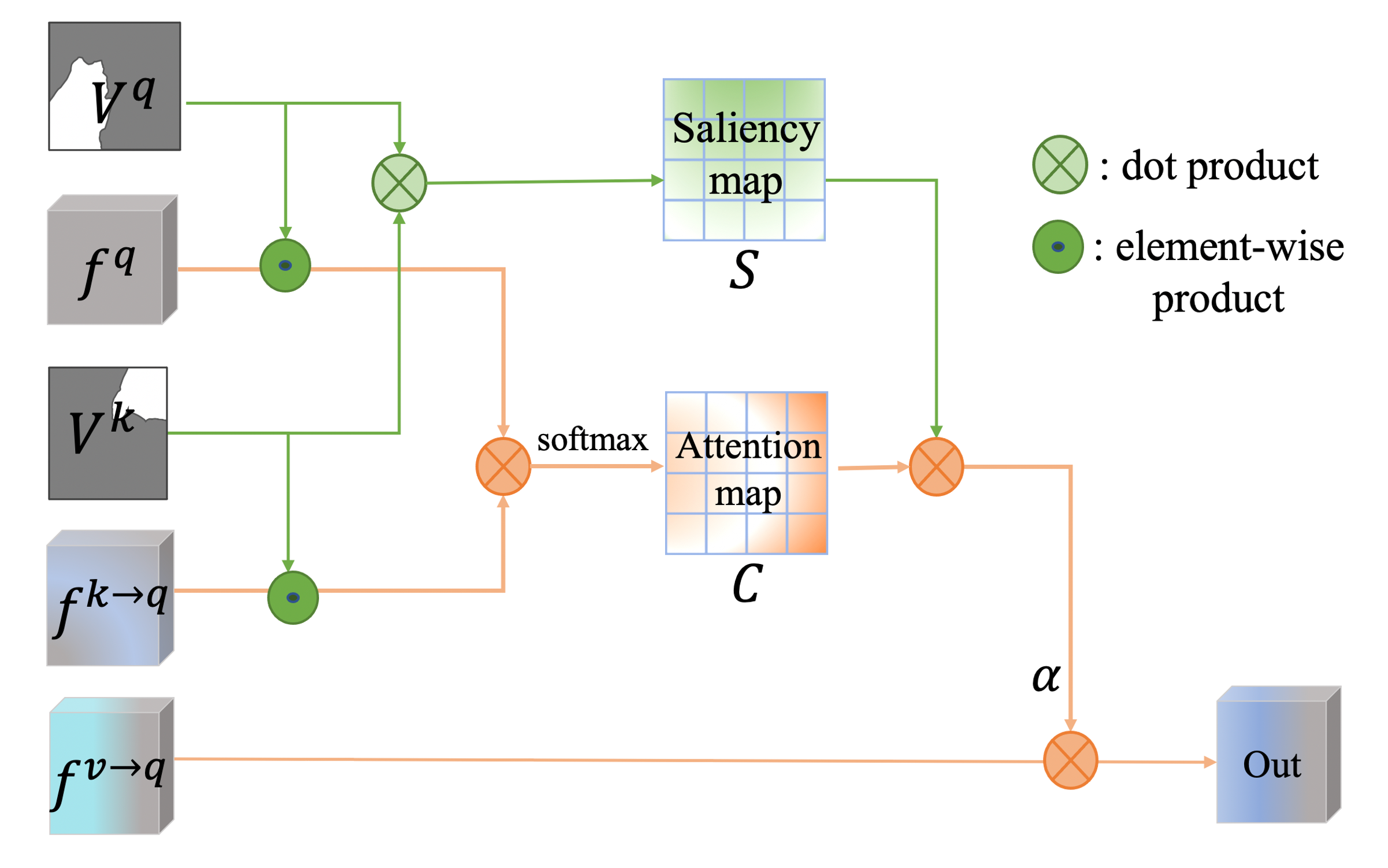}
    \caption{Mask pruning based Patch attention(MPPA). MPPA pruned out invalid pixels confusing the matching similarity in warped tokens, thus the attention mechanism is more engaging to video inpainting task.}
  \end{center}
\label{fig:MPPA}
\vspace{-4mm}
\end{figure}

\subsection{Pruned Patch Attention in Transformer}
\label{sec:3.3}
Invalid pixels (holes) in token prevent the model from generating smooth results. The traditional Patch Matching mechanism takes no account of the structural characteristics of two patches. However, the confidence for matching scores should be reduced when the large invalid area of two patches overlaps since these invalid pixels will generate blurry results although they have a similar structure.
With this in mind, we propose Mask Pruning-based Patch Attention (MPPA) mechanism, aiming to achieve a better elaborate matching among the tokens containing invalid pixels. First, MPPA prunes out invalid pixels from the warped query and key tokens and then calculates the correlation among them. Second, the final score is weighted by the correlation map according to the valid common ratio of two patches (saliency map). Our new mechanism (MPPA) could choose the most contributing features from the pruned features and reconstructs the final feature map for the decoder more accurately. MPPA can be regarded as a generalization of attention operators in traditional transformers. When there is no invalid pixel in the patch, MPPA degrades into conventional attention operation in the transformer.

\textbf{Mask-Pruning based Attention (MPPA)} 
MPPA is based on patch triplet, whose patch matching produces a measure of similarity between the pruned Q, K, V patch feature. We denote the basic operation between the query patch and aligned key patch as $\boldsymbol{O}\left(\boldsymbol{f}^{k}, \boldsymbol{f}^{q}\right):=\boldsymbol{C}$, such that $\boldsymbol{C}:\mathds{R}^{N \times d} \times \mathds{R}^{d \times N} \rightarrow \mathds{R}^{N \times N}$, where $N=T \times N_{p}$ is the number of patches in T frames, $d=c \times h \times w$ is the hidden feature of the patch. $\boldsymbol{C}$ is used to compute the similarity between pruned patch pair (pruned query patch $\boldsymbol{f}^{q}$ and pruned aligned key patch $\boldsymbol{f}^{k \rightarrow q}$)\footnote{The symbol $\rightarrow$ in superscript indicates alignment. $\boldsymbol{f}^{k \rightarrow q}$ indicates key patch $k$ aligned to query patch $q$. $\boldsymbol{V}^{k \rightarrow q}$ indicates the visibility map of $k$ aligned to $q$.}. We constrain the matching to happen only between the valid parts excluding the hole pixels. To this end, we use down sampled binary inpainting masks $m^{k}$ and $m^{q} \in \mathds{R}^{N \times N}$. With $i$ and $j$ denoting the patch indices for $\boldsymbol{f}^{k}$ and $\boldsymbol{f}^{q}$ respectively, the correlation map is computed. For a pair of Query patch and Key patch, we compute pruned patch similarities $\boldsymbol{C}(q, k)$ between Query patch $q$ and aligned Key patch $k$ in the feature space:
\begin{equation}
\label{eq:1}
\boldsymbol{C}(q, k)= \sum_{(x, y)} \boldsymbol{V}^{q}(x, y) \cdot \boldsymbol{f}^{q}(x, y) \cdot  \boldsymbol{V}^{k \rightarrow q}(x, y) \cdot \boldsymbol{f}^{k \rightarrow q}(x, y)
\end{equation}

where $\boldsymbol{V}=1-m$ is the valid map, 
$x$, $y$ are pixels in valid areas. $\boldsymbol{V}^{k \rightarrow q}$ is the aligned valid map of key. The above equation is basically computing the pruned similarity between two aligned patches, excluding the hole pixels. Then a saliency map $S(q,k)$ for each pair of patch is computed as follows:
\begin{equation}
\label{eq:2}
    S(q, k)= \frac{1}{\sum \boldsymbol{V}} \cdot \sum_{(x, y)} \boldsymbol{V}^{k \rightarrow q}(x, y) \cdot \boldsymbol{V}^{q}(x, y)
\end{equation}

Each pixel value in the saliency map (correlation map) holds the weight that specific pixels have on filling the hole in the target. The attention map is the similarities $\boldsymbol{C}$ weighting on the saliency map.
\begin{equation}
\label{eq:3}
Attn(q, k) = \boldsymbol{C}(q, k) \cdot S(q, k)
\end{equation}
\begin{equation}
\label{eq:4}
\alpha_{q, k} = softmax(Attn(q, k))
\end{equation}
Finally, the aligned value features are aggregated through a weight sum with the attention map, and the output of each patch can be obtained by weighted summation of aligned value patch features $\boldsymbol{f}^{v\rightarrow q}$ from relevant patches:
\begin{equation}
\label{eq:5}
    MPPA(\boldsymbol{f}^{q}, \boldsymbol{f}^{k\rightarrow q},  \boldsymbol{f}^{v\rightarrow q}) = \alpha_{q, k} \cdot \boldsymbol{f}^{v\rightarrow q}
\end{equation}
Examples of calculating MPPA are given in supplementary materials. The MPPA mechanism follows the concept of the multi-headed self-attention in transformer technique while in accordance with the video inpainting task. 

\begin{figure}[t]
  \centering
  \includegraphics[width=0.45\textwidth]{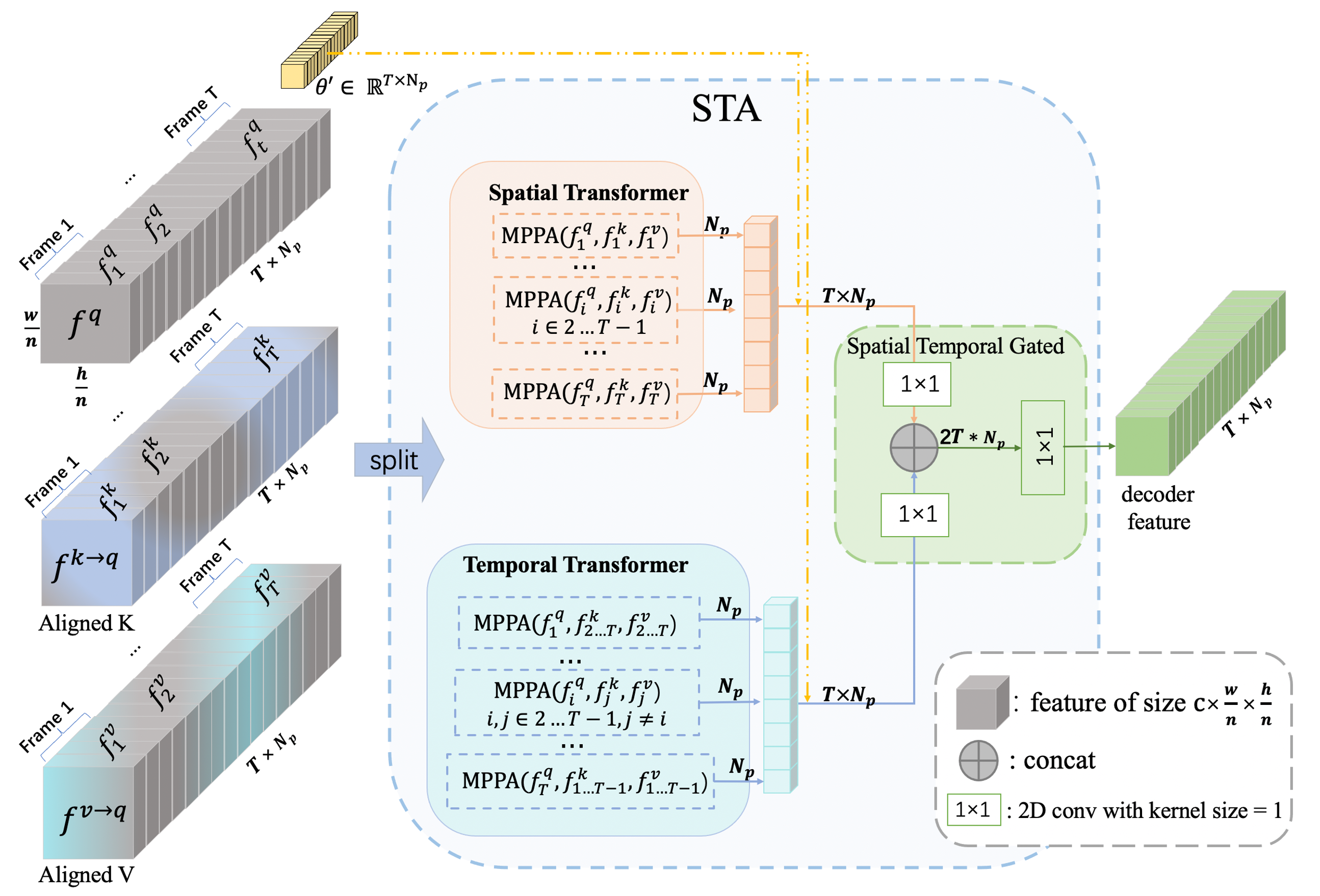}
  \caption{\label{fig:STA} Spatial-temporal weighting Adaptor (STA). The Query (Q), Aligned Key (K), and Aligned Value (V) are sliced according to patches in different frames, and perform attention operation (MPPA) (annotated with square dotted line) across patches {\bf in} frames and {\bf cross} frames. $N_{p} = n^2$ is number of patches in each frames while different transformer head has different $N_{p}$, and $n=2, N_{p}=4$ in this case. The orange block is the spatial branch, and the blue block is the temporal branch. STA does not increase computational complexity compared to vanilla transformers since they compute equal size attention maps.}
  \vspace{-4mm}
\end{figure}


\begin{figure*}[t]
  \centering
  \includegraphics[width=0.95\linewidth]{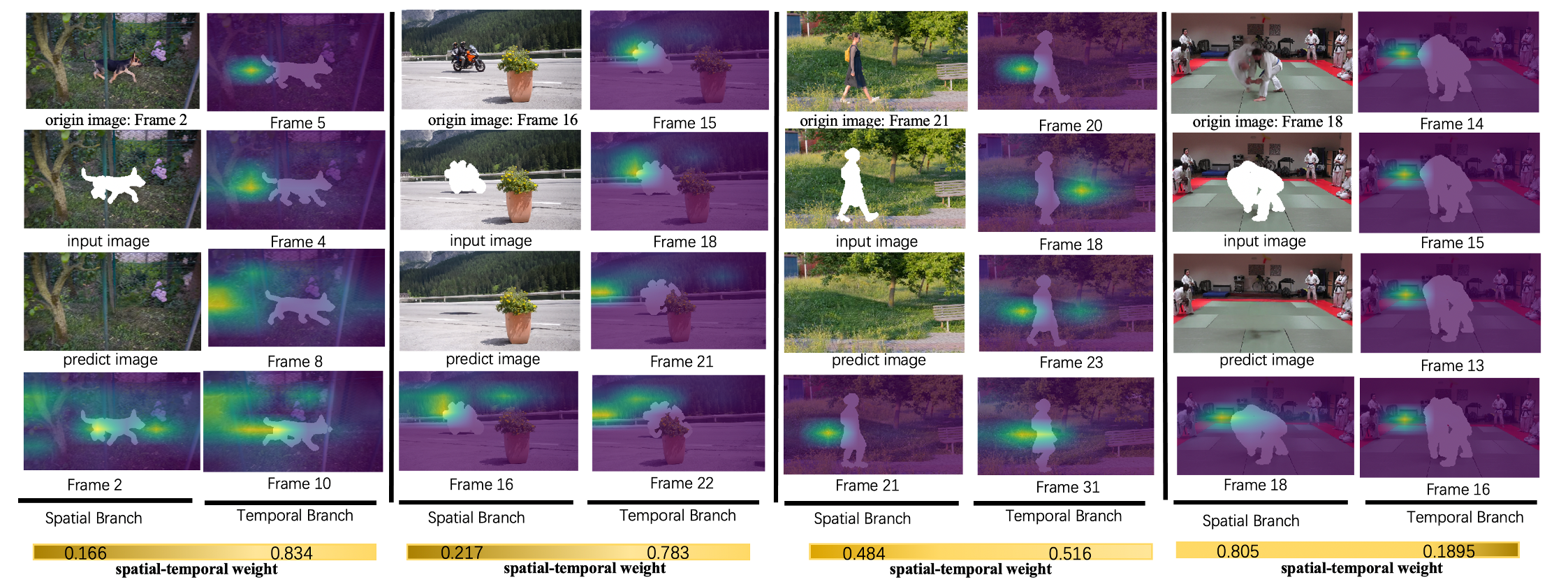}
   \caption{\label{fig:attention maps} The attention maps learned by our model. Four cases from left to right correspond to motion types C, B, B, and A. STA assigns different spatial-temporal weights to videos with different motion types. Our model can track the moving dog and motorbike over the video mainly by attending to temporal tokens under the guidance of Deformation Factor $\theta'$. For completing the human mask with local motion, STA attends to the spatial token. Thus the target frame itself owns the highest attention score. Attention regions are highlighted in yellow. The number in the final line indicates the attention weight for the spatial and temporal branches.}
   \label{fig:onecol}
\end{figure*}

\vspace{-4mm}
\subsection{Spatial Temporal Weighting Adaptor (STA)}
 Different types of videos need reference patches in different locations, which own different spatial or temporal effectiveness. Specifically, on the one hand, spatial patches are applicable to many scenarios, such as non-repetitive scene motion or under the hand-held camera condition. On the other hand, temporal patches are mainly used to handle moving scenes. However, some previous methods ~\cite{Free-form} only use spatial information or pay equal attention to spatial and temporal patches for different scenarios ~\cite{STTN}, which is not adaptive for complex and varied scenes. 
 To overcome these difficulties, we propose a weighting adaptor that can compute spatial-temporal weight according to learned video characteristics ($\theta$ learned by DePtH could model the moving range of object motion). By assigning different spatial-temporal weights to patches in different branches, different kinds of movement would be modeled by paying attention to spatial contents or variant motion differently.

Figure \ref{fig:STA} shows the Spatial-Temporal weighting Adaptor in our network. The orange block is the spatial branch, and the blue block is the temporal branch. The input of our STA is the query (Q), deformed key (K), and deformed value (V) output from DePtH, as well as a deformation factor $\theta'$ learned by DePtH as guidance. The three features will be sliced in patch level and perform pruned-based attention operation (MPPA) (annotated with a square dotted line) across patches inside each frame and between different frames. 

We conduct MPPA for patches of different sizes in the different heads. In practice, we first extract spatial patches of shape $N_{p} \times c \times \frac{w}{n} \times \frac{h}{n}$ from the query feature of each frame, and obtain $N_{p}=n^2$ patched query feature for each frame. Then we do spatial and temporal transformer for patches in these $T$ frames: 
\begin{equation}
    \small
    \label{eq:6}
    T_{m,n}=MPPA(\boldsymbol{p}_{\boldsymbol{m,i}}^{\boldsymbol{q}}, \boldsymbol{p}_{\boldsymbol{n,j}}^{\boldsymbol{k}\rightarrow \boldsymbol{q}}, \boldsymbol{p}_{\boldsymbol{n,j}}^{\boldsymbol{v}\rightarrow \boldsymbol{q}})
\end{equation}

where $1 \leq i$, $j \leq N_p$, $1 \leq m$, $n \leq T$. $w_{p}$ and $h_{p}$ are the shape of each patches. $\boldsymbol{p}_{m,i}^{q}$denotes the $i$-th query patch in $m$-th frame, $\boldsymbol{p}_{n,j}^{k \rightarrow q}$ denotes the $j$-th query patch in $n$-th frame. When $m=n$, the Spatial Transformer is conducted, as shown in orange block in the attention map. When $m\ne n$, Temporal Transformer is conducted, as shown in blue block in the attention map. The STA does not increase computational complexity compared to traditional transformer since they compute equal size attention map, which are visually presented to demonstrate our computation.
We compute $T$ times spatial attention which produce $N_{p} \times N_{p}$ attention map each time, and $T$ times temporal attention which produce $N_{p} \times (\mathrm{T}-1) N_{p}$ attention map each time. The output of spatial branch will be concatenated to feature of shape $N_{p} \times t \times c \times \frac{w}{n} \times \frac{h}{n}$, and add with motion encoding $\theta^{\prime}$ which is learned by DePtH representing the motion pattern of the video. The same procedure is applied to the Temporal Transformer branch. These two branches will be fused by Spatial-Temporal Gate as illustrated in Figure~\ref{fig:STA} under the guidance of $\theta$' learned by DePtH. 
Figure~\ref{fig:attention maps} reveals the attention map of our STA module. Since the deformation factor $\theta'$ implicitly reveals the dynamic pattern of the video, STA can allocate more accurate weights to spatial/temporal tokens under the guidance of the learned deformation factor $\theta'$. In the first case in Figure~\ref{fig:attention maps}, for stable foreground-background motions(the dog), the model could attend to the correct location of patches in neighbor frames. In the second case(the water ripple), for the video with local motion, the learned deformation factor $\theta'$ guided STA to attend to the local patches since the top 1 attended image is the target image inpainted. In conclusion, STA learns the different spatial-temporal weights to handle videos with different moving speeds uniformly and enhance the temporal consistency in repaired areas. 


\vspace{-0.1in}
\subsection{Training Objective}
We end-to-end optimize our DeViT in a self-supervised manner. Choosing optimization objectives ensures per-pixel reconstruction accuracy, perceptual rationality, and spatial-temporal coherence in generated videos~\cite{Free-form,gatys2016image,johnson2016perceptual,Copy-and-paste}. To this end, we select a pixel-wise reconstruction loss focuses on the pixel-level features. 
\begin{equation}
\vspace{-1mm}
    \small
    \label{eq:8}
    \mathcal{L}_{\text{hole}}=\sum_{j}\left\|H^{j} \odot\left(\hat{Y}^{j}-Y\right)\right\|_{1}, 
    \mathcal{L}_{\text{valid }}=\sum_{j}\left\|V \odot\left(\hat{Y}^{j}-Y\right)\right\|_{1}
\vspace{-1mm}
\end{equation}
Where H is hole pixels and V are valid pixels. Inspired by the recent studies~\cite{Free-form}, we use adversarial training to help our model generate high-quality content generation results. Specifically, we use a Temporal PatchGAN (T-PatchGAN) as our discriminator~\cite{Free-form,chang2019learnable,yang2020learning,zeng2019learning}. In order to enhance both perceptual quality and spatial-temporal coherence in video inpainting ~\cite{Free-form,chang2019learnable}. This spatial-temporal adversarial loss has shown promising results. The T-PatchGAN discriminator is composed of 6 3D convolutional layers with kernel size 3 × 5 × 5 and stride
1×2×2. The recently proposed spectral normalization ~\cite{miyato2018spectral}
is applied to both the generator and discriminator, similar to
~\cite{nazeri2019edgeconnect} to enhance training stability. In addition, we use the hinge loss as the objective function to discriminate if the input video is real or fake.

\begin{figure*}[t]
  \centering
  \includegraphics[width=0.95\textwidth]{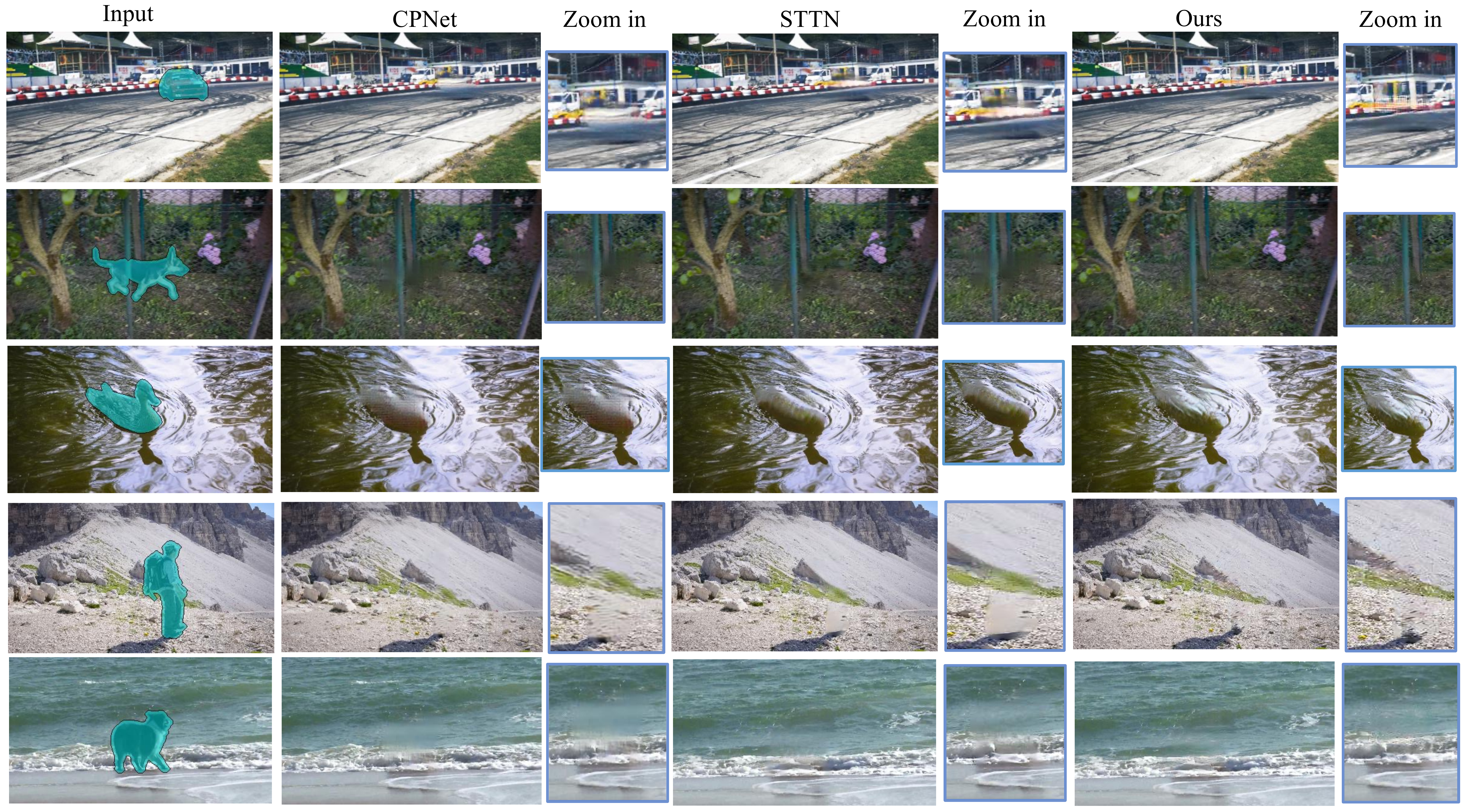}
  \caption{\label{fig:compare}Qualitative comparison with other methods on DAVIS for object removal. The 1st and 2nd row are cases of slow panning motions(type B), mainly using a temporal token. The 3rd to 5th row are cases of local motion (type A) which needs more spatial token. Our method accurately finds the correct position of patches in both motion types.}
  \vspace{-4mm}
\end{figure*}

The optimization function for the T-PatchGAN \cite{Free-form} discriminator is shown as follows:
\begin{equation}
\vspace{-2mm}
\label{eq:9}
\begin{aligned}
    L_{D}&=E_{x \sim P_{Y_{1}^{T}}(x)}[\operatorname{ReLU}(1-D(x))] \\
    &+E_{z \sim P_{\hat{\Upsilon}_{1}^{T}}(z)}[\operatorname{ReLU}(1+D(z))]
\end{aligned}
\vspace{-2mm}
\end{equation}

The adversarial loss is denoted as: 
\vspace{-3mm}
\begin{equation}
\vspace{-2mm}
\label{eq:10}
    L_{adv}=-E_{z \sim P_{\hat{Y}_{1}^{T}}(z)}[D(z)]
\vspace{-2mm}
\end{equation}


The total loss is the weighted summation of all the loss functions:
\begin{equation}
\vspace{-2mm}
\label{eq:11}
\mathcal{L}_{\text{total}} =  \lambda_{hole} \cdot \mathcal{L}_{\text{hole}} + \lambda_{val} \cdot \mathcal{L}_{\text{valid}}
+ \lambda_{adv} \cdot \mathcal{L}_{\text{adv}}
\vspace{-2mm}
\end{equation}

\section{Experiments}
\label{experiments}
\subsection{Datasets}
We use DAVIS~\cite{perazzi2016benchmark} and DAVIS~\cite{lin2019tsm, liu2018image} for train and evaluation. 
We split DAVIS~\cite{perazzi2016benchmark} dataset into a training set including 90 video clips and a test set including 60 video clips. To simulate real-world applications, we synthesized datasets for two tasks: video completion and object removal. We evaluate models for the two tasks using two types of free-form masks, including stationary masks and moving masks~\cite{chang2019learnable, kim2019deep, Copy-and-paste}. For the video completion task, a sequence of stationary masks is generated to simulate applications like watermark removal, hoping the algorithm can recover the original videos. For the object removal task, we synthesized videos by putting imaginary object masks from DAVIS~\cite{lin2019tsm, liu2018image} and Youtube-VOS~\cite{nazeri2019edgeconnect} on the videos. The original video is used as the ground truth. \label{sec:4.2}

\vspace{-4mm}
\subsection{Experimental Analysis}
 To simulate real-world applications, we perform experiments by comparing our network with other deep video inpainting approaches to video completion and object removal. For testing video completion tasks, the original videos are taken as ground truths, and we evaluate models from both quantitative and qualitative aspects. For testing moving masks, we use foreground object annotations as masks since the ground truths after foreground removal are unavailable. We evaluate the models through qualitative analysis following previous works. We split the video into three motion types: \textbf{A}: no motion, \textbf{B}: slow panning motions, \textbf{C}: agile/complex motions.
 
 \begin{figure*}[ht]
  \centering
  \includegraphics[width=0.96\textwidth]{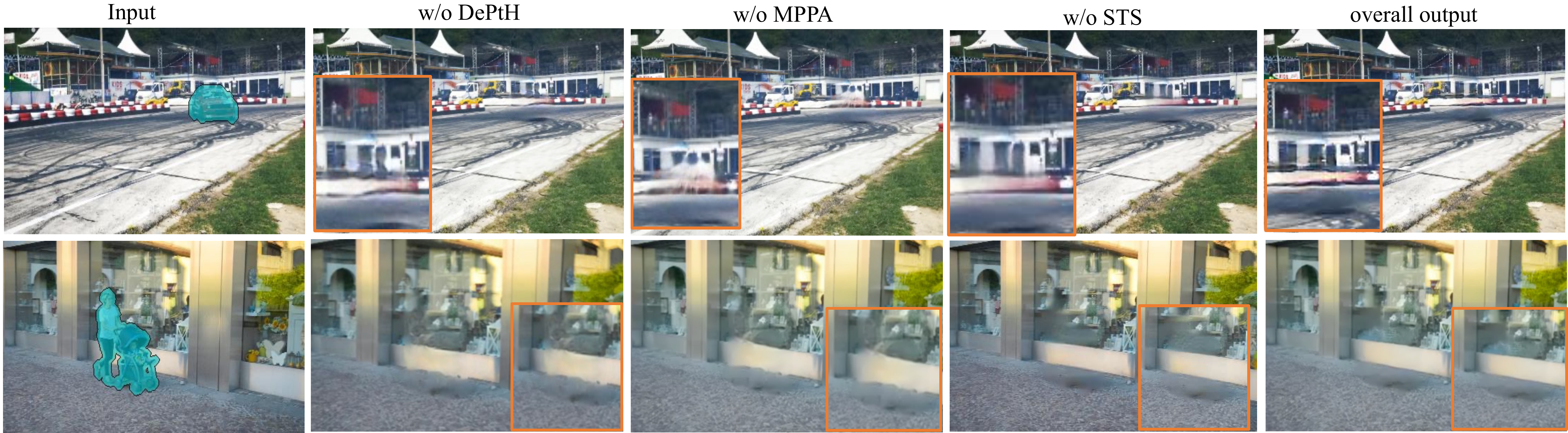}
  \vspace{-0.1in}
  \caption{\label{fig:ablation}Ablative comparison. We show the result without(w/o) DePtH, MPPA, and STA, respectively. DePtH helps to correct the unwarp lines by better dealing with scaled and rotated patches, which is robust to inter-patch misalignment. MPPA could avoid mistakenly patch retrieval. STA could localize more accurate patches from spacial/temporal contents according to the motion type of video. The ensemble result hence enjoys better quality and robustness.}
\vspace{-3mm}
\end{figure*}

\begin{table}
	\footnotesize
	\setlength\tabcolsep{1.5pt}
	\centering
	\caption{The quantitative evaluation of video inpainting. We compare our method with 8 different baselines under four different metrics. Our model achieves state-of-the-art results.}
	\vspace{-3mm}
	\label{tab:inpaintinglabel} 
	\begin{center}
		\scalebox{1}{
		\begin{tabular}{c|cccc|cccc}
			\toprule
			{\multirow{2}{*}{Models}}  &\multicolumn{4}{c}{YouTube VOS} &\multicolumn{4}{c}{DAVIS}
			\\
			\cmidrule{2-9}
             & PSNR & SSIM & VFID & $E_{warp}$ & PSNR & SSIM & VFID & $E_{warp}$
            \\
			\midrule
			VINet~\cite{kim2019deep}& 29.20 & 0.9434 & 0.1490 & 0.072 & 28.96 & 0.9411 & 0.1785 & 0.199
			\\
			DFVI~\cite{Deep-flow-guided} & 29.16 & 0.9429 & 0.1509 & 0.066 & 28.81 & 0.9404 & 0.1880 & 0.187 
			\\
			CPT~\cite{Copy-and-paste} & 31.58 & 0.9607  & 0.1470& 0.071 & 30.28  & 0.9521 & 0.1824 & 0.182  
			\\
			FGVC~\cite{gao2020flow} & 31.83 & 0.9634 & 0.1460 & 0.062 & 30.51  & 0.9535 & 0.1791 & 0.165
			\\
			STTN~\cite{STTN}& 32.34 & 0.9655  & 0.1451 & 0.053 & 30.67 & 0.9560 & 0.1779 & 0.149
			\\
			Decouple~\cite{liu2021decoupled}& 32.66 & 0.9646  & 0.1430 & 0.052 & 31.75 & 0.9650 & 0.1716 & 0.148
			\\
			Fuseformer~\cite{liu2021fuseformer}& 33.16 & 0.9673  & - & 0.051 & \textbf{32.54} & 0.9700 & - & 0.138
			\\
			TSAM~\cite{zou2021progressive}& 32.71 & 0.9713 & 0.1433 & 0.053 & 31.77 & 0.9566 & 0.1731 & 0.141
			\\
			Ours & \textbf{33.42}& \textbf{0.9732} & \textbf{0.1429} & \textbf{0.049} & 32.43 & \textbf{0.9721} & \textbf{0.1663} & \textbf{0.133}
			\\
			\bottomrule
		\end{tabular}
		}
	\end{center}
\vspace{-6mm}
\end{table}

\textbf{Qualitative Analyses}
As shown in Figure~\ref{fig:compare}, our model achieves competitive qualitative performance against other alignment methods~\cite{Copy-and-paste}, flow guided methods~\cite{Deep-flow-guided,gao2020flow,zou2021progressive,kim2019deep}, and Transformer based methods~\cite{STTN,liu2021decoupled,liu2021fuseformer}. CPT cannot deal with camera and local motions such as water ripples. Transformer based methods~\cite{STTN,liu2021decoupled,liu2021fuseformer} yield blurry results when dealing with agile/complex motions. Flow-based approaches~\cite{Deep-flow-guided,zou2021progressive} are sensitive to errors in optical flow and are able to generate higher resolution results for video of motion type A. However, Our model outperforms both methods in any motion type in either motion scenes or stationary scenes and has large gains on agile motions(type \textbf{C}). 
 

\begin{table}
	\footnotesize
	\setlength\tabcolsep{1.5pt}
	\centering
	\caption{\label{tab: Ablation1} Ablation study on the effect of different module. w/o means without. We remove some of our blocks to verify the effectiveness of them. $STA_{s}$ is the spatial branch of $STA$ and $STA_{t}$ is the temporal branch of $STA$.}
	\label{tab:distribution_sim}
	\vspace{-3mm}
	\begin{center}
		\scalebox{1.0}{
		\begin{tabular}{c|cc|cc}
			\toprule
			{\multirow{2}{*}{Models}}  &\multicolumn{2}{c}{YouTube VOS} &\multicolumn{2}{c}{DAVIS}
			\\
			\cmidrule{2-5}
             & PSNR & SSIM  & PSNR & SSIM
            \\
			\midrule
			Transformer baseline & 31.62 & 0.9573 & 30.64 & 0.9555
			\\
			w/o $DePtH$ & 31.85 & 0.9647 & 31.95 & 0.9621   
			\\
			w/o $STA_{s}$ & 32.55 & 0.9679 & 32.29 & 0.9662  
			\\
			w/o $STA_{t}$ & 32.03 & 0.9582 & 32.11 & 0.9643
			\\
			w/o $\theta^{'}$ & 33.13 & 0.9743 & 32.38 & 0.9716
			\\
			overall & \textbf{33.42}& \textbf{0.9732} & \textbf{32.43} & \textbf{0.9721}
			\\
			\bottomrule
		\end{tabular}
		}
	\end{center}
	\vspace{-4mm}
\end{table}

\begin{figure*}[t]
\vspace{-3mm}
  \begin{center}
    \includegraphics[width=0.95\textwidth]{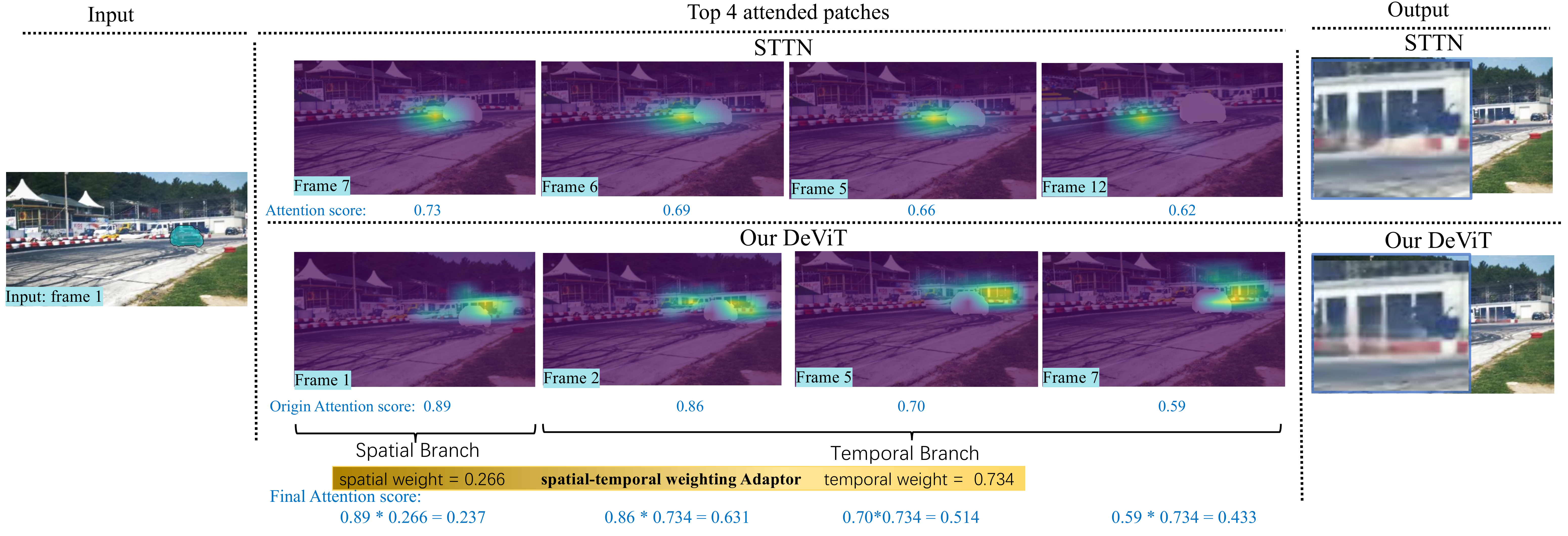}
    \vspace{-0.1in}
    \caption{\label{fig9} An example for reference token attention. We compared our DeViT against conventional Transformer~\cite{STTN}. DePtH and MPPA help to attend to more accurate localization in each frame. Moreover, the motion type is taken into account in our DeViT; thus, STA computes the final attention score by giving a higher rank to more accurate attended frames. Our result has fewer twisted lines compared with STTN~\cite{STTN} method.}
  \end{center}
  \vspace{-4mm}
\end{figure*}

\begin{table}
	\footnotesize
	\setlength\tabcolsep{1.5pt}
	\centering
	\caption{\label{tab: Ablation2} Ablation study on the effect of MPPA. w/o means without. We remove some of our operation to verify the effectiveness of each one of them.}
	\label{tab:distribution_sim1}
	\vspace{-2mm}
	\begin{center}
		\scalebox{1.0}{
		\begin{tabular}{c|cc|cc}
			\toprule
			{\multirow{2}{*}{Models}}  &\multicolumn{2}{c}{YouTube VOS} &\multicolumn{2}{c}{DAVIS}
			\\
			\cmidrule{2-5}
             & PSNR & SSIM  & PSNR & SSIM
            \\
			\midrule
			w/o MPPA & 32.21 & 0.9603 & 31.31  & 0.9563
			\\
			w/o mask pruning & 32.68 & 0.9625 & 31.67 & 0.9571  
			\\
			w/o saliency map & 33.04 & 0.9681 & 31.90 & 0.9639
			\\
			overall & \textbf{33.42}& \textbf{0.9732} & \textbf{32.43} & \textbf{0.9721}
			\\
			\bottomrule
		\end{tabular}
		}
	\end{center}
	\vspace{-4mm}
\end{table}

\textbf{Quantitative Analyses} We conducted a quantitative evaluation by measuring the quality of video completion and object removal. We use PSNR, SSIM, optical flow-based warping error ($E_{warp}$)~\cite{lai2018learning}, and Video-based \text{Fréchet Inception Distance} (VFID)~\cite{wang2018video} as they are the most widely-used metrics for video quality assessment. Table~\ref{tab:inpaintinglabel} compares the PSNR and the SSIM measures between our method and other state-of-the-art methods. Our results are the best among the compared methods. Finally, we also conducted a user study on the visual quality of different methods in supplementary material.

\vspace{-2mm}
\subsection{Ablation Study}
 We perform an ablation study to investigate how each component in our network impacts the performance of video completion. The quantitative metrics PSNR and SSIM at 200 000 iterations are summarized in Table~\ref{tab: Ablation1}. We investigate \textbf{A},\textbf{B},\textbf{C} three motion types, and choose 30 videos from DAVIS validate set (10 type A, 10 type B and 10 type C). We also randomly choose 60 videos (20 for each type) from the YouTube-VOS test set. our sub-module are compared against other methods~\cite{liu2021fuseformer,loshchilov2018decoupled,gao2020flow,zou2021progressive} for each type. 

\begin{table}
	\footnotesize
	\setlength\tabcolsep{1.5pt}
	\centering
	\caption{\label{tab: Ablation1} Ablation study different motion types. `vanilla' means Transformer baseline. DePtH gains more on type B than type C and STA gains more on type B/C than A.} 
	\label{tab:distribution_sim}
	\vspace{-3mm}
	\begin{center}
		\scalebox{1.0}{
		\begin{tabular}{c|ccc|ccc}
			\toprule
			{\multirow{3}{*}{PSNR}}
			&\multicolumn{3}{c}{YouTube VOS}
			&\multicolumn{3}{c}{DAVIS}
			\\
            \cmidrule{2-7}
             & \textbf{A} (33\%) & \textbf{B} (33\%)& \textbf{C} (33\%) & \textbf{A}(33\%)  & \textbf{B}(33\%) & \textbf{C}(33\%)
            \\
			\midrule
			vanilla & 31.99 & 32.65 & 31.34& 31.10 & 32.22 & 31.27
			\\
			$decouple$~\cite{liu2021decoupled}& 32.31 & 32.71 & 32.35 & 32.34  & 32.82 & 31.57
			\\
			FGVC~\cite{gao2020flow} & 32.21 & 32.58  & 31.93 & 31.50  & 32.64  & 31.21
			\\
			vanilla+$F3N$~\cite{liu2021fuseformer}& 32.24 & 33.09  & 31.62 & 32.21  & 32.45 & 31.32
			\\
			$Fuseformer$~\cite{liu2021fuseformer}& 32.34 & 32.62  & 31.84 & \textbf{32.44}  & 32.35 & 31.43
			\\
			vanilla+$STA$ & 32.15 & 33.65 & 31.63 & 32.09 & 32.51 & 31.72
			\\
			vanilla+$DePtH$ & 32.43 & 33.63 & 31.99 & 32.01 & 32.84 & 31.89
			\\
			TSAM~\cite{zou2021progressive}& 32.61 & 33.43  & 32.47 & 32.23  & 32.88 & 31.39
			\\
			ours & \textbf{32.81} & \textbf{33.83} & \textbf{32.84} &  32.39  & \textbf{33.05} & \textbf{32.12}
			\\
			\bottomrule
		\end{tabular}
		}
	\end{center}
	\vspace{-5mm}
\end{table}

\textbf{The Effect of DePtH} DePtH helps to obtain accurate content that is more structure-preserving by warping patch-based features in the attention mechanism. As shown in Figure~\ref{fig:ablation}, when we remove the DePtH module (the first column), the model is non-robust to inter-patch misalignment (scaled and rotated patches), and the straight lines are twisted as the attended patches (V) did not warp to the accurate angle compared to the target patch (Q).

\textbf{The Effect of MPPA} MPPA aims to accurately localize the target patch to fill in the hole by eliminating the participation of the hole pixel in the matching score calculation. Without MPPA, the model mistakenly attends to the patches with holes (on the mask's boundary), which typically leads to incorrect results.

\textbf{The Effect of STA} In the Youtube-VOS dataset, many videos own a stationary background. Thus, STA will allocate more weight to the spatial branch. As shown in Figure~\ref{fig:ablation}, without STA, the inpainted textures yield blurry outputs for mistakenly attending to irrelevant patches. As shown in Fig~\ref{fig9}, STA can adaptively learn spatial and temporal weight for different motion types. 

\textbf{The Effect on Different Motion Types} The performance across different motion types is in Table~\ref{tab: Ablation1}. We compare all methods under the same settings and mask type. The quantitative results show that DePtH gains more on type B than C, and STA gains more on type B/C than A. The qualitative results are shown in Fig~\ref{fig:attention maps}.

\textbf{Training Details}
\label{training-details} 
 We use $240 \times 432$ images for training. We use Adam optimizer with $\beta_{1}=0.99, \beta_{2}=0.999$ and its learning rate starts from $1e^{-4}$ and is decayed once with the factor 0.1 at 150,000 iteration. The total training iteration is 500,000. We use two 2080Ti GPUs for training and set the batch size to 2. The proposed DeViT can enable fast training and inference. Specifically, our model runs at about 17.1 fps with an NVIDIA 2080Ti GPU. Our full model has 28.8M trainable parameters and 266 G FLOPS. It consumes about 3.9G GPU memory for completing a video from the DAVIS dataset, which is the same order of magnitude with~\cite{STTN}. More ablation studies and failure case analyses are in the supplementary.


\section{Conclusion}
\label{Conclusion}

\vspace{-2mm}
In this paper, we proposed a novel framework DeViT with emphasis on better patch-wise alignment and matching in video inpainting. Specifically, a new DePtH is designed to introduce alignment in the patch token. Aligned tokens are then adaptively selected by MPPA, which could improve feature matching quality. Moreover, the STA in the transformer balances spatial and temporal branches for better adaptation to different scenarios (motion types). Extensive experiments verify the effectiveness of our proposed modules. Our method achieves state-of-the-art performance for video inpainting.

\textbf{Acknowledgement:} This work was supported by SZSTC Grant No.JCYJ20190809172201639 and WDZC20200820200655001, Shenzhen Key Laboratory ZDSYS20210623092001004.

\clearpage
\bibliographystyle{ACM-Reference-Format}
\balance
\bibliography{egbib}


\clearpage

\appendix
\section{appendix}
\subsection{Details of Network Architecture}
\label{1}
\textbf{Deformed Transformation estimator} Table~\ref{tab: Transformer} illustrate the detailed network architectures of our Deformed Transformation estimator module described in Sec~\ref{Method} of the main paper. All the convolutional layers are followed by ReLU for the non-linearity except for the last layer of decoder.

Transformation estimator takes the correlation map C as input and produces homography parameters $\theta$ between the reference and target. It is trained to output 6 parameters (i.e.affine transformation), such that $R(C) =\theta$, and $R: \mathds{R}^{N\times c \times wh} \rightarrow \mathds{R}^{N \times 2\times3}$.

\textbf{Temporal Patch GAN} 
The DeViT is built on a Generative Adversarial Network(GAN). Inspired by \cite{Free-form}'s work, we integrate the temporal dimension and use Temporal PathGAN (T-PatchGAN) as our discriminator that focuses on different spatial-temporal features to fully utilize all the global and local image features and temporal information together. T-PatchGAN discriminator is composed of six 3D convolutional layers with kernel size 3 $\times$ 5 $\times$ 5 and stride 1 $\times$ 2 $\times$ 2. 3d-Conv denotes a 3D convolution layer that adopts spectral normalization to stabilize GAN's training. The recently proposed spectral normalization is applied to both the generator and discriminator, similar to~\cite{Free-form} to enhance training stability. Similar with~\cite{Free-form}, we use the hinge loss as the objective function as to discriminate if the input video is real or fake. The details of the architectures of T-PatchGAN is in table \ref{tab: T-PatchGAN}.





\begin{figure}[h]
 \vspace{-4mm}
  \begin{center}
    \includegraphics[width=0.4\textwidth]{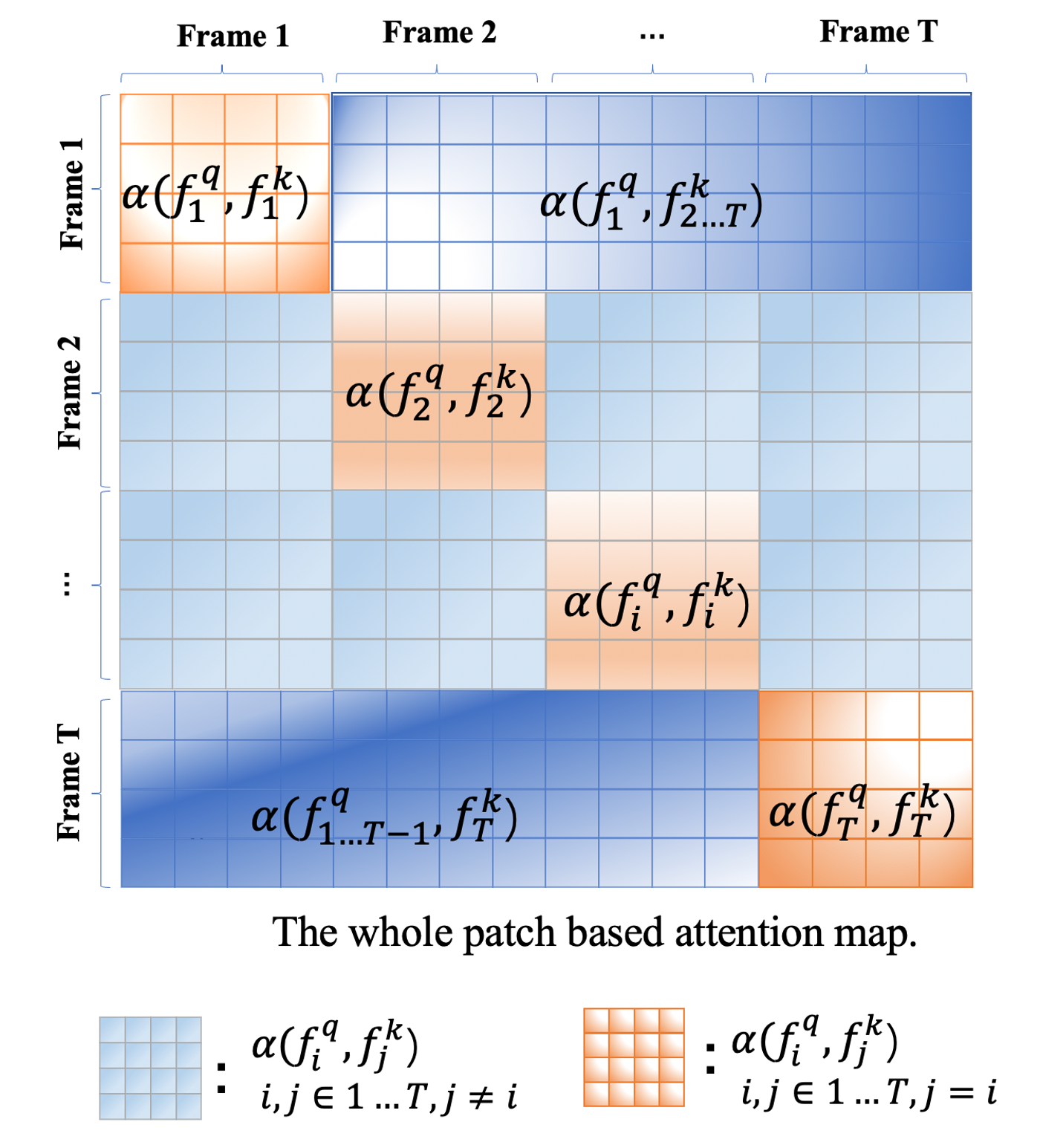}
    \caption{\label{suppfig3} The whole patch-based attention map.}
  \end{center}
   \vspace{-4mm}
\end{figure}

\textbf{Spatial Temporal Branches} Figure~\ref{suppfig3} demonstrate the visualization attention map for whole patches. When $i=j$, the Spatial Transformer is conducted, as shown in orange block in the attention map. When $i \ne j$, Temporal Transformer is conducted, as shown in blue block in the attention map. The spatial temporal weighting adaptor (STA) does not increase computational complexity compared to conventional transformer since they compute equal size attention map, which are visually presented to demonstrate our computation. We compute $T$ times spatial attention which produce $N_{p} \times N_{p}$ attention map each time, and $T$ times temporal attention which produce $N_{p} \times(\mathrm{T}-1) N_{p}$ attention map each time. The output of spatial branch will be concat to feature of shape $N_{p} \times t \times c \times \frac{w}{n} \times \frac{h}{n}$, and add with motion encoding $\theta^{\prime}$ which is learned by DePtH representing the motion pattern of the video. The same procedure is applied to the Temporal Transformer branch. These two branches will be fused by Spatial Temporal Gate as illustrated in Figure 3 in main paper under the guidance of $\theta$ learned by DePtH.
 
\subsection{Computation complexity}
We focus on the computation complexity caused by the MPPA in spatial and temporal branches and leave out other computation costs (e.g., encoding and decoding costs). The computation complexity of the proposed DeViT with Spatial-Temporal weighting Adaptor (STA) are denoted as:
\begin{small}
\begin{align*}
\small
\mathcal{O}\left(n^{2}\right) &  \approx \mathcal{O}\left(\sum_{l=1}^{D}\left[ \left( \frac{H W}{N_{p}}\right)^{2} \cdot\left({N_{p}} C_{l}\right)+n k_{l}^{2} H W C_{l-1} C_{l}\right]\right)  \\
& + \mathcal{O}\left(\sum_{l=1}^{D} \left[  n \cdot \left( \frac{H W}  {N_{p}} \right) \cdot  \left( n-1 \right) \cdot \left(\frac{H W}  {N_{p}} \right) \cdot  \left({N_{p}} C_{l} \right) +n k_{l}^{2} H W C_{l-1} C_{l}\right]\right)
\end{align*}
\end{small}

$N_{p}$ is the number of patches in each frame, n is the number of input frames, $HW$ is the feature size, $k_{l}$ denotes for kernel size, $C$ is the channel number of features and $D$ is number of convolutional layers. We demonstrate that the Spatial-Temporal weighting Adaptor (STA) and Mask Pruning based Patch Attention (MPPA) doesn't increase computational complexity compared to traditional transformer since they compute equal size attention map. 


The Spatial-temporal Selective mechanism can also used as a plugin, which can be inserted into transformers and replace the conventional Attention computation to increase the spatial-temporal selection when dealing with variance video datasets.

\begin{table}[h]
	\footnotesize
	\setlength\tabcolsep{5pt}
	\centering
	\caption{\label{tab: Transformer} Details of the Transformer architecture.}
	\vspace{-4mm}
	\begin{center}
		\scalebox{0.8}{
		\begin{tabular}{c|c|c|c|c}
            \text { Module Name } & \text { Filter Size } & \text { Channels } & \text { Stride $/$ Up Factor } & \text { Nonlinearity } \\
            \hline \text { 2dConv } & 3 $\times$ 3 & 64 & 2 & \text { LeakyReLU(0.2) } \\
            \text { 2dConv } & 3 $\times$ 3 & 64 & 1 & \text { LeakyReLU(0.2) } \\
            \text { 2dConv } & 3 $\times$ 3 & 128 & 2 & \text { LeakyReLU(0.2) } \\
            \text { 2dConv } & 3 $\times$ 3 & 256 & 1 & \text { LeakyReLU(0.2) } \\
            \hline \text { Transformer } $\times$ 8 & 1 $\times$ 1 & 256 & 1 & $-$ \\
            & 3 $\times$ 3 & & 1 & \text { LeakyReLU(0.2) } \\
            \hline \text { BilinearUpSample } & - & 256 & 2 & $-$ \\
            \text { 2dConv } & 3 $\times$ 3 & 128 & 1 & \text { LeakyReLU(0.2) } \\
            \text { 2dConv } & 3 $\times$ 3 & 64 & 1 & \text { LeakyReLU(0.2) } \\
            \hline \text { BilinearUpSample } & $-$ & 64 & 2 & $-$ \\
            \text { 2dConv } & 3 $\times$ 3 & 64 & 1 & \text { LeakyReLU(0.2) } \\
            \text { 2dConv } & 3 $\times$ 3 & 3 & 1 & \text { Tanh }

        \end{tabular}}
	\vspace{-4mm}
	\end{center}
\end{table}

\begin{table}[h]
	\footnotesize
	\setlength\tabcolsep{5pt}
	\centering
	\caption{\label{tab: T-PatchGAN} Details of the T-PatchGAN discriminator.}
	\label{tab:distribution_sim}
	\vspace{-4mm}
	\begin{center}
		\scalebox{0.75}{
		\begin{tabular}{c|c|c|c|c|c}
            \text { Module Name } & \text { Filter Size } & \text { Channels } & \text { Stride } & \text { Nonlinearity } & \text{normalize}\\
            \hline  \text { 3d-Conv } & 3 $\times$ 5 $\times$ 5 & 64 & (1,2,2) & \text { LeakyReLU(0.2) } & \text { spectral norm}\\
            \text { 3d-Conv } & 3 $\times$ 5 $\times$ 5 & 128 & (1,2,2) & \text { LeakyReLU(0.2) } & \text { spectral norm}\\
            \text { 3d-Conv } & 3 $\times$ 5 $\times$ 5 & 256 & (1,2,2) & \text { LeakyReLU(0.2) }& \text { spectral norm} \\
            \text { 3d-Conv } & 3 $\times$ 5 $\times$ 5 & 256 & (1,2,2) & \text { LeakyReLU(0.2) } & \text { spectral norm}\\
            \text { 3d-Conv } & 3 $\times$ 5 $\times$ 5 & 256 & (1,2,2) & \text { LeakyReLU(0.2) } & \text { spectral norm}\\
            \text { 3d-Conv } & 3 $\times$ 5 $\times$ 5 & 256 & (1,2,2) & $-$ & \text { spectral norm}\\ 
            \hline
        \end{tabular}
        }
	\vspace{-4mm}
	\end{center}
\end{table}



\begin{figure*}[h]
  \begin{center}
    \vspace{-4mm}
    \includegraphics[width=0.9\textwidth]{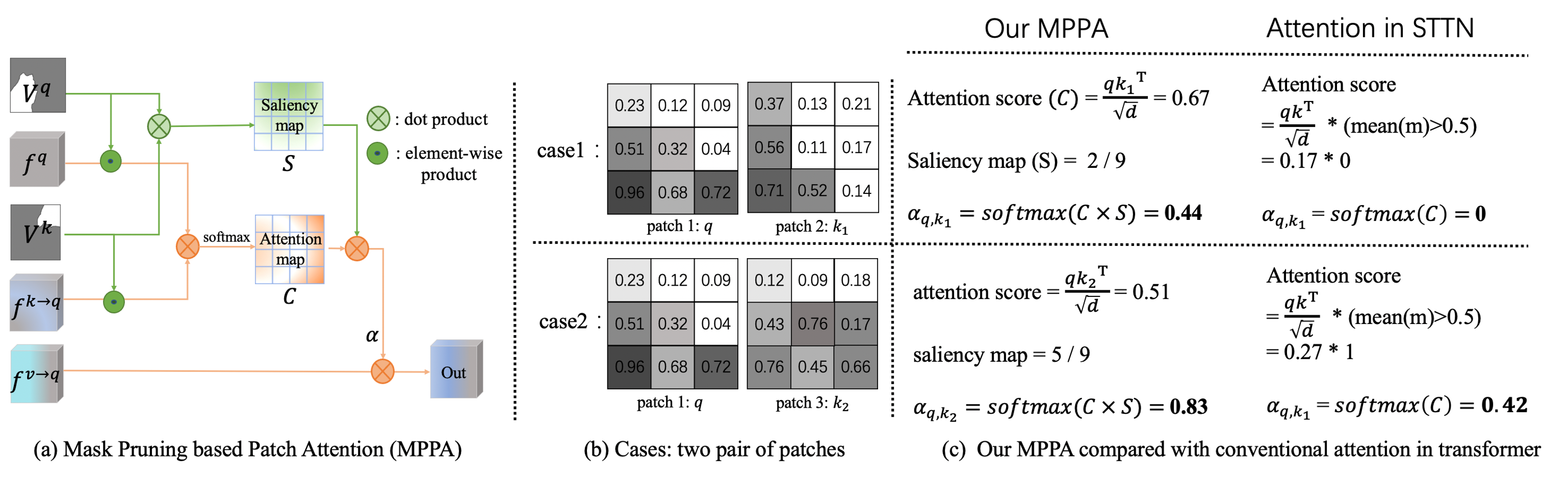}
    \caption{\label{fig:suppfig4} An example of calculating our MPPA versus conventional attention operation.}
    \vspace{-4mm}
  \end{center}

\end{figure*}

\begin{figure*}[h]
  \begin{center}
    \includegraphics[width=0.9\textwidth]{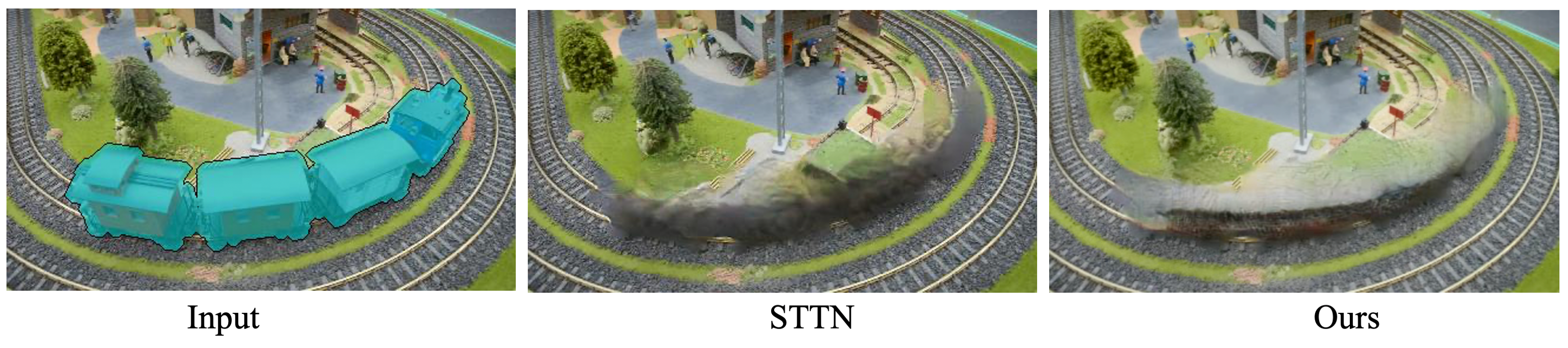}
    \vspace{-4mm}
    \caption{\label{fig:failure} A failure case. The third column shows our results with large holes. For reconstructing the train occluded by a large mask, although better than STTN, DeVit fails to find enough patches to refer to and generates blurs inside the mask.}
  \end{center}
  \vspace{-4mm}
\end{figure*}

\begin{figure}
    \includegraphics[width=0.5\textwidth]{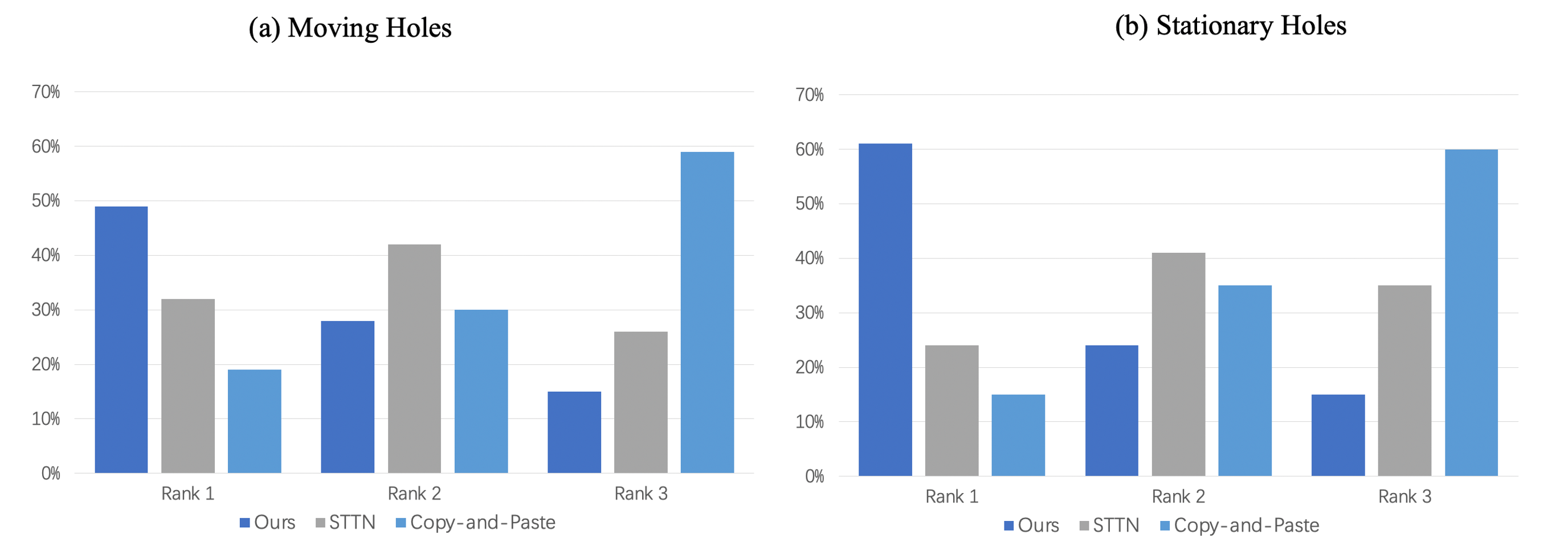}
    \caption{\label{fig:userstudy} User study on different methods. We compare our method with STTN~\cite{STTN} and Copy-and-Paste~\cite{Copy-and-paste}. Our model is ranked in first place in most cases.}
\end{figure}

\subsection{More Experiments}
\label{3}
\textbf{User Study} We conduct a user study for a more comprehensive comparison. we choose STTN~\cite{STTN} and Copy-ans-Paste~\cite{Copy-and-paste} as two baselines, since we have observed their significantly better performance than other baselines from both quantitative and qualitative results. We randomly sampled 20 videos (10 from DAVIS~\cite{perazzi2016benchmark} and 10 from Youtube-VOS~\cite{nazeri2019edgeconnect}) for stationary masks filling, and 20 videos from DAVIS for moving masks filling. In practice, 20 volunteers are invited to the user study. In each trial, inpainting results from different models are shown to the volunteers, and the volunteers are required to rank the inpainting results. The results of the user study are concluded in Figure~\ref{fig:userstudy}. We can find that our model performs better in most cases for these two types of masks.

\textbf{Cases of MPPA} MPPA aims at accurately localise the target patch to fill in the hole by eliminating participation of hole pixel in the matching score calculation. Without MPPA, the model mistakenly attend to the wrong patches, which typically leads to incorrect results. Figure~\ref{fig:suppfig4} demonstrate two pair of patches that compute attention score using our MPPA versus conventional attention in STTN. (a) demonstrate our MPPA architecture, (b) is two pair of query (q) and key (k), which number in each grid is the normalized feature of each pixel. 

\textbf{Effectiveness of mask pruning} In case 1, our MPPA compute the attention score of $q$ and $k_{1}$ according to the section 3.3, which result is 0.44. However, in STTN, the conventional attention score is 0.17, and the weight with the ratio of hole pixel in $k_{1}'s$ mask which $mean(m)<0.5$. So the final attention weight is 0. In other words, although patch pair $q$ and $k_{1}$ share some same information, STTN delete this key for the reason that more than half of the pixel in $k_{1}$ is in the hole, since conventional attention in STTN cannot capture information from $k{1}$ to query patch, and prevented the process from locating high quality semantically similar candidates.

\textbf{Effectiveness of saliency map} our MPPA compute the saliency map respectively for case 1 and case 2. According to the equation 2 in section 3.3, the saliency map (valid weighing map) holds the weight that valid pixels have on filling the hole in the target. The saliency score for case 1 is $\frac{2}{9}$ and $\frac{5}{9}$ for case 2. We hope the model pay more attention to the patch pair whose common valid pixels ratio is higher.

\subsection{Limitations And Societal Impact}
\label{4}
\textbf{Limitations} We note that DeViT may generate blurs in large missing masks if continuous quick motions occur. As shown in Figure~\ref{fig:failure}, although our DeViT performs better than transformer baseline STTN~\cite{STTN}, DeViT generates blurs when reconstructing the large hole. We infer that when the hole is very large, there are no abundant patches for DeViT to calculate attention scores. We will investigate other types of temporal losses for joint optimization in the future.

\textbf{Societal Impact} Malicious use of DeViT may bring negative societal impacts: fake videos could be used to deceive or persuade viewers on social media. We believe that the task itself is neutral with positive uses, such as image/video editing in TV and movie post-production, to improve storytelling and expression.

\end{document}